\documentclass{midl} 

\usepackage{mwe} 
\usepackage{multirow}

\usepackage{array}  

\usepackage[symbol]{footmisc}

\usepackage{caption} 

\title[Label Noise-Resilient Histopathology Image Classification]{Contrastive-Based Deep Embeddings for Label Noise-Resilient Histopathology Image Classification}

\midlauthor{\Name{Lucas Dedieu\nametag{$^{1}$}} \Email{lucas@primaalab.com}\\
\Name{Nicolas Nerrienet\nametag{$^{1}$}} \Email{nicolas.n@primaalab.com}\\
\Name{Adrien Nivaggioli\nametag{$^{1}$}} \Email{adrien@primaalab.com}\\
\Name{Clara Simmat\nametag{$^{1}$}} \Email{clara@primaalab.com}\\
\Name{Marceau Clavel\nametag{$^{1}$}} \Email{marceau@primaalab.com}\\
\Name{Arnaud Gauthier\nametag{$^{1,2}$}} \Email{arnaud@primaalab.com}\\
\Name{Stéphane Sockeel\nametag{$^{1}$}} \Email{stephane@primaalab.com}\\
\Name{Rémy Peyret\nametag{$^{1}$}} \Email{remy@primaalab.com}\\
\addr $^{1}$ Primaa, Paris, France \\
\addr $^{2}$ Diagnostic and Theranostic Medicine Division, Department of Pathology, Institut Curie, PSL University, Paris, France\\
}

\begin{document}

\maketitle

\begin{abstract}
Recent advancements in deep learning have proven highly effective in medical image classification, notably within histopathology. However, noisy labels represent a critical challenge in histopathology image classification, where accurate annotations are vital for training robust deep learning models. Indeed, deep neural networks can easily overfit label noise, leading to severe degradations in model performance. While numerous public pathology foundation models have emerged recently, none have evaluated their resilience to label noise. Through thorough empirical analyses across multiple datasets, we exhibit the label noise resilience property of embeddings extracted from foundation models trained in a self-supervised contrastive manner. We demonstrate that training with such embeddings substantially enhances label noise robustness when compared to non-contrastive-based ones as well as commonly used noise-resilient methods. Our results unequivocally underline the superiority of contrastive learning in effectively mitigating the label noise challenge. Code is publicly available at \url{https://github.com/LucasDedieu/NoiseResilientHistopathology}
\end{abstract}


\section{Introduction}
Histopathology, the microscopic examination of tissue samples, plays a crucial role in the diagnosis, prognosis, and treatment of various diseases, including cancer. With the rise of digital pathology thanks to whole-slide image (WSI) scanners, image analysis and deep learning are becoming part of histopathologists' routine. As the workload of histopathologists has grown significantly in previous decades~\cite{garcia}, deep learning-based image analysis now plays a major role in increasing clinical workflow efficiency~\cite{dawson}. To be effective, training such deep neural networks (DNNs) requires large image datasets with reliable labels. However, in the context of medical imaging and histopathology in particular, clean data are rare and expensive, requiring expert labeling campaigns. Thus, one of the key challenges in histopathology image analysis is the presence of label noise, which refers to errors or inaccuracies in the annotations provided for the training data. These inaccuracies, stemming from inter-observer variability, imperfect segmentation of tissue regions, inherent ambiguity in the biological features, and omission errors, impede the development of reliable deep learning models. Indeed, it has been proven that DNNs can easily overfit noisy labels~\cite{Li2018LearningTL, Zhang2016UnderstandingDL}, leading to severe degradations in model performance and thus potentially misleading clinical decisions. 

To mitigate this challenge, various approaches have been proposed in the literature, ranging from robust loss functions to novel training methodologies and label cleaning strategies. In more recent works, self-supervised-based methods also emerged. Nevertheless, the quest for noise-tolerant models in the presence of label noise remains an ongoing pursuit.

In this paper, we aim to comprehensively evaluate the robustness of deep embeddings extracted from pretrained backbones specifically designed for histopathology image classification. We compare the noise tolerance of models trained solely on these features to those trained on the original images and assess their performance on multiple histopathology datasets. Our key contributions are: (1) We demonstrate that classifiers trained on contrastive deep embeddings exhibit improved robustness to label noise compared to those trained on the original images using state-of-the-art methods. This highlights the inherent noise resilience of these features. (2) We compare the noise robustness of contrastive embeddings to alternative self-supervised embeddings extracted from non-contrastive backbones, demonstrating the superior ability of contrastive-based backbones to handle noisy labels.

\section{Related Work}

\subsection{Learning with Noisy Labels Methods}
In image classification, a multitude of approaches have been explored to enhance the robustness of learning algorithms in the presence of noisy labels.


\paragraph{Label Cleaning Strategies:}
 CleanNet \cite{cleannet} uses a small, clean dataset to generate feature vectors for each class, comparing them with query image vectors to determine label accuracy. Rank Pruning \cite{rankpruning}  focuses on high-confidence data points for classifier training, potentially overlooking challenging samples common in medical tasks. To overcome this, \citet{ehn} propose a two-phase hard sample aware noise robust learning algorithm consisting in an easy/hard/noisy detection scheme combined with a noise supressing and hard enhancing scheme.

\paragraph{Robust Loss Functions:}
Other methods focus on changing the loss function, such as \citet{ghosh} that showed that the categorical cross entropy (CCE) loss is sensitive to label noise while the mean absolute error (MAE) loss is robust. However, training a network under MAE loss leads to underfitting and thus \citet{gce} proposed generalized cross entropy (GCE), a noise-robust loss function that can be seen as a generalization of MAE and CCE. In the same way, \citet{APL} showed that every loss can be noise-robust through simple normalization, but may lead to an underfitting problem. To tackle this challenge, the authors introduce a synergistic boosting framework that enables the development of a novel family of loss functions, each possessing the theoretical guarantee of robustness while maintaining efficient learning.

\paragraph{Training Procedures:}
Numerous studies also explored novel training approaches to address the issue of label noise. Co-teaching \cite{coteaching} proposed a new deep learning paradigm that involves maintaining two neural networks in parallel and selecting instances with low loss values for cross-training. Improving this work, Co-teaching+ \cite{coteaching+} bridges "update by disagreement" strategy with the original Co-teaching. Also using two networks, DivideMix \cite{dividemix} employs a mixture model to dynamically partition the training data into clean and noisy subsets, enabling semi-supervised learning on both. To address confirmation bias, the two networks are trained separately, each using the other network's dataset division.

\subsection{Self-Supervised-Based Methods}
Some recent works leverage advances in self-supervised learning (SSL) to deal with noisy labels. SSL is a paradigm in which models are trained from unlabeled data using pretext tasks. By solving these pretext tasks, models learn meaningful representations of the data and the acquired knowledge is then transferred to the primary task of classification or segmentation. Tackling label noise, Co-learning \cite{colearning} trains two classifier heads with a shared feature encoder and regularizes the model with both the intrinsic similarity and the structural similarity. \citet{kurian} use a contrastive learning framework and feature aggregating memory banks to identify and increase the emphasis on clean training samples. \citet{selfattention} integrate contrastive learning and intra-group attention mixup techniques into standard supervised learning. \citet{onlyssl} showed that model pretraining with SimCLR \cite{simclr} consistently yields the best results in presence of noisy labels on two medical datasets. 

The reasons behind the enhancement of label noise resilience through contrastive learning are still not fully understood, and as far as our knowledge extends, only one study \cite{theorie} has delved into the theoretical understanding of this achievement. The authors demonstrated that contrastive learning yields a representation matrix characterized by a significant gap between the prominent singular values and the remaining ones, along with a considerable alignment between the prominent singular vectors and the accurate labels. Aforementioned characteristics enable a linear classifier head trained on these representations to adeptly learn clean labels while minimizing overfitting to the noisy ones.

\subsection{Pathology Foundation Models}
The emergence of SSL has been particularly impactful in the domain of histology, where obtaining large amounts of accurately labeled data is challenging and expensive. In the last few years, many foundation models trained in a self-supervised way have been developed. \citet{ctranspath,ctranspath2} introduced CTransPath, a feature extractor based on the swin transformer \cite{swintrans}, trained on TCGA \cite{tcga} and PAIP \cite{paip} using semantically-relevant contrastive learning (SRCL), a novel SSL technique adapted from MoCo v3 \cite{mocov3} specifically designed for pathology applications. They also presented RetCCL \cite{retccl}, a ResNet-50 \cite{resnet} model trained on the same two datasets using their clustering-guided contrastive learning (CCL) SSL technique, which was based on MoCo \cite{moco}. \citet{phikon} evaluated various vision transformer (ViT) variants \cite{vit} on TCGA using the iBOT framework \cite{ibot} and mask image modeling, a new paradigm of SSL coming from the world of natural language processing (NLP) applied to vision. Authors named their most effective ViT-B variant "Phikon". Lunit \cite{lunit} conducted a benchmark of several SSL techniques, including Barlow Twins \cite{barlowtwin}, SwAV \cite{swav}, MoCo v2 \cite{mocov2}, and DINO \cite{dino} (self-distillation), for pathology by training them on TCGA. \citet{pathoduet} trained Pathoduet on TCGA using newly-introduced pretext tokens and later task raisers to explicitly utilize certain relations between images and showed better results than CTransPath on some patch classification tasks.
In recent months, several other pathology foundation models have emerged, trained on much larger datasets \cite{uni, virchow, rudolfv, 3billions, conch}. Regrettably, we could not incorporate these models into our research due to their proprietary nature.


\section{Proposed Approach}

\subsection{Problem Formulation}
Given a \(K\)-class image classification problem, let \(\mathcal{X} \subset \mathbb{R}^d\) be the image space, and let \(\mathcal{Y}=\{1,...,K\}\) be the class labels. In a typical learning scheme, we are given training data, \(S=\{(\mathbf{x},y)^{(i)}\}_{i=1}^{N} \in (\mathcal{X}\times \mathcal{Y})^{N}\). Classification aims to learn a function \(f : \mathcal{X} \rightarrow \mathcal{Y}\) (represented by a DNN) that maps the input space to the label space. Training model \(f\) involves finding optimal parameters \(\theta\) minimizing the empirical risk, i.e., \(\theta := \arg \min_\theta \sum_{i=1}^{N} \mathcal{L}(f(\mathbf{x_i}), y_i))\), where \(\mathcal{L}(f(\mathbf{x}), y)\) is the loss of \(f\) with respect to \(y\).

In presence of uniform label noise, the noisy training data available to the classifier is \(S_\eta=\{(\mathbf{x},\hat{y})^{(i)}\}_{i=1}^{N}\), where \(\eta \in [0, 1]\) denotes the uniform noise rate in the dataset, and
\begin{equation}
    \hat{y}_i =\left\{ \begin{array}{rcl}
            y_i & \mbox{with probability} & (1-\eta) \\ 
            k , k\neq y_i& \mbox{with probability} & p_\eta=\frac{\eta}{K-1} \\
            \end{array}\right.
\end{equation}

\begin{remark}
We do not consider a uniform noise rate \(\eta > \frac{K-1}{K}\) as, for balanced dataset, it would mean that for each class \(k\), $\exists c\neq k,$ where $c$ has more data labeled as class $k$ than $k$.
\end{remark}

In asymmetric label noise scenarios, the flipping probabilities of each class are conditioned by a transition matrix $T$, thus $p_\eta =  T_{y_i,k}$ represents the probability of transitioning from true label $y_i$ to noisy label $k$. Asymmetric label noise tends to provide a more realistic representation of real-world noise scenarios as certain classes may be more prone to confusion due to visual similarity or contextual ambiguity. 

\subsection{Learning with Contrastive-Based 
 Deep Embeddings}
To mitigate label noise, we propose an approach based on contrastive-based deep embeddings. We employ a contrastive-pretrained backbone to extract embeddings from images and then exploit these embeddings and corresponding noisy labels to train a linear classifier head. Here, $f = h\circ g$ with $g : \mathcal{X}\rightarrow\mathcal{B}$, the frozen backbone parameterized by $\theta_g$ and $h : \mathcal{B}\rightarrow\mathcal{Y}$, the classifier head parameterized  by $\theta_h$. $\mathcal{B}$ represents the latent space of embeddings. The main advantage of this training approach is that it allows to use the same pre-trained feature extractor $g$ for multiple datasets, and only train a separate classifier $h_i$ for each dataset $i=1,...,M$. It enables significantly faster trainings since the latent space $\mathcal{B}$ and parameters $\theta_h$ are much smaller than the image space $\mathcal{X}$ and $\theta$.

\section{Experiments}
In this section, we propose a comparison of different methods for handling noisy labels. We benchmark four image-based methods: a  baseline, GCE, Active-Passive loss (APL), and DivideMix. In addition, we compare eight deep embedding methods based on eight foundation models. These include four contrastive histology backbones: RetCCL, CTransPath, PathoDuet, and Lunit Barlow Twins (BT); two non-contrastive histology backbones: Phikon and Lunit DINO; and two ViT-B models trained on ImageNet \cite{imagenet}  with MoCo v3 (contrastive) and iBOT (non-contrastive). We include the two ImageNet models to also investigate the performance of contrastive and non-contrastive embeddings derived from backbones in unrelated domains. The experimental settings are detailed in appendix \ref{settings}.

\subsection{Datasets and Pre-processing}
We validate the effectiveness of our proposed approach by assessing its performance on six public benchmark datasets, namely NCT-CRC-HE-100K \cite{crc}, PatchCamelyon \cite{pcam}, BACH \cite{bach}, MHIST \cite{mhist}, LC25000 \cite{lc25000} and GasHisSDB \cite{gashis}. \tableref{tab:dataset}, summarizes all dataset details. We add more information about these datasets and the pre-processing used in appendix \ref{dataset}.

\begin{table}[h]
    \scriptsize
    \centering
    \caption[]{Dataset details. \footnote[2] : set obtained from train set split, \footnote[3] : set labellized by a pathologist expert for this study.}
        {\setlength{\tabcolsep}{1em}
        \begin{tabular}{||c|c|c|c|c|c||}
            \hline
            Dataset & Patch Size & Train & Validation & Test & $K$\\
            \hline
            \hline
            NCT-CRC-HE-100k & 224x224 & 80,000 & 20,000\footnote[2] & 7,180 & 9\\
            PatchCamelyon & 96x96 & 262,144 & 32,768 & 32,768 & 2 \\
            BACH  & 2048x1536 & 320 & 80\footnote[2] & 83\footnote[3] & 4\\
            MHIST & 224x224 & 1740 & 435\footnote[2] & 977 &2\\
            LC25000 & 768x768 & 16,000 & 4000\footnote[2] & 5000\footnote[2] & 5\\
            GasHisSDB &  160x160 & 21,303& 5325\footnote[2] & 6656\footnote[2] &2\\

            \hline
        \end{tabular}
        }
        
\label{tab:dataset}
\end{table}


\subsection{Results}
\begin{figure}[h]
\centering
\includegraphics[width=14cm]{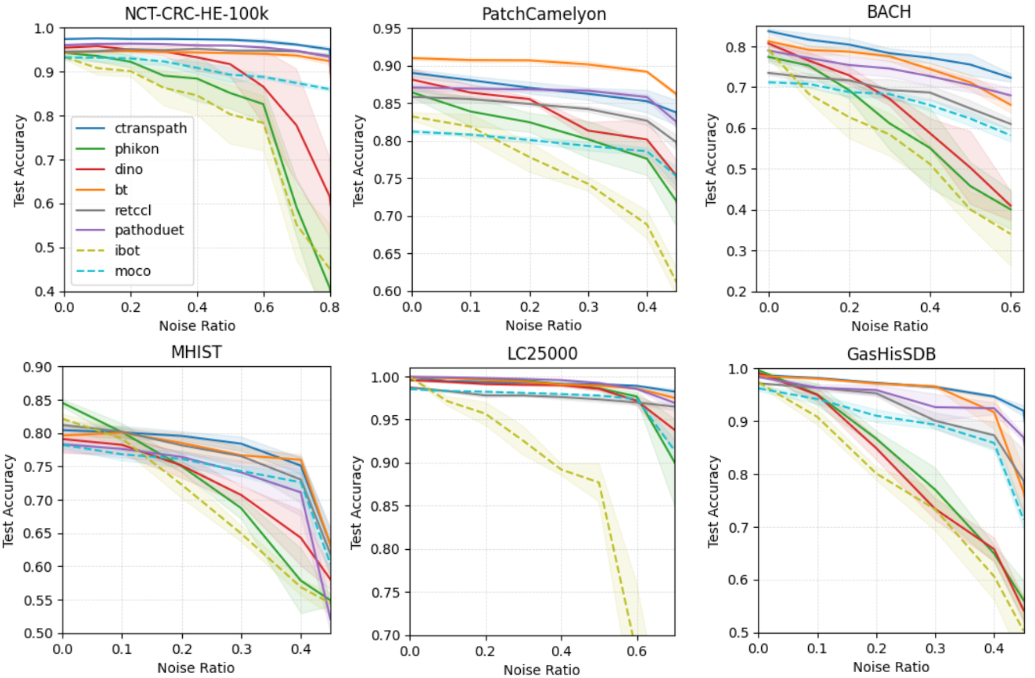}
\caption{Average test accuracies (4 runs) of linear classifiers trained with deep embedding over different label noise ratios. Shaded areas represent standard deviation. iBOT and MoCo backbones (dashed curves) are pre-trained on ImageNet.}
\label{fig:curves}
\end{figure}

The results under uniform label noise, as depicted in \tableref{tab:sym}, highlight the efficacy of training with deep embeddings. Across nearly all the datasets and noise rates scenarios, these methods consistently match or surpass performances of image-based approaches. Particularly noteworthy is the observation that, for noise rates $\eta > 0$, classifiers trained with contrastive embeddings exhibit superior performance compared to their non-contrastive counterparts. This phenomenon is graphically illustrated in \figureref{fig:curves}, providing a finer-grained insight into the impact of noise rates. Across all datasets, the accuracies of Phikon and Lunit-DINO (non-contrastive) consistently exhibit sharper declines compared to others. 
\newcolumntype{C}[1]{>{\centering\arraybackslash}p{#1}}
\begin{table}[h]
    \scriptsize
    \centering
    \caption[]{Average test accuracies with standard deviation under different uniform label noise ratios (\%, 2 runs for image methods, 4 runs for deep embedding methods). \footnote[1] :Foundation model trained with contrastive learning.}
    
    \begin{tabular}{||C{1.7cm}||C{1cm}C{1cm}C{1cm}C{1cm}C{1.2cm}|C{1cm}C{1cm}C{1.2cm}||}
        \hline
        \multirow{2}{*}{Method} & \multicolumn{5}{c|}{NCT-CRC-HE-100k} & \multicolumn{3}{c||}{PatchCamelyon}\\
        & $\eta=0$ & $\eta=0.2$ & $\eta=0.4$ & $\eta=0.6$ & $\eta=0.8$ & 
        $\eta=0$ &  $\eta=0.2$ & $\eta=0.4$ \\
        \hline
        \hline
        Baseline & 96.7$\pm$0.1 & 90.4$\pm$0.7 & 83.2$\pm$1.7 & 70.9$\pm$6.1 & 35.5$\pm$0.3 & 87.3$\pm$2.1 &85.9$\pm$1.0 & 73.1$\pm$1.8 \\
        GCE & 96.8$\pm$0.2 & 96.1$\pm$0.5 & 94.8$\pm$0.4 & 93.4$\pm$0.2 & 82.9$\pm$5.3 & 85.5$\pm$1.1 & 85.2$\pm$1.6 & 82.8$\pm$1.0 \\
        APL & 96.4$\pm$0.1 & 96.6$\pm$0.1 & 96.2$\pm$0.1 & 95.1$\pm$0.2 & 8.8$\pm$0.1 & 89.1$\pm$0.1 & 86.4$\pm$0.4 & 81.2$\pm$3.3 \\
        DivideMix & 97.1$\pm$0.1 & 96.9$\pm$0.1 & 96.4$\pm$0.2 & 96.3$\pm$0.2 & 94.7$\pm$1.7 & 90.3$\pm$0.1 & 87.4$\pm$0.2 & 87.0$\pm$0.7 \\
        \hline
        Phikon & 94.3$\pm$0.6 & 92.2$\pm$1.4 & 88.4$\pm$5.2 & 82.6$\pm$4.0 & 40.9$\pm$11.2 & 86.4$\pm$0.7& 82.5$\pm$1.3 & 77.6$\pm$2.2 \\
        RetCCL\footnote[1] & 94.5$\pm$0.1 & 95.1$\pm$0.2 & 95.2$\pm$0.1 & 94.8$\pm$0.2 & 93.6$\pm$0.3 & 85.8$\pm$0.7 & 84.5$\pm$0.2 & 82.7$\pm$0.6\\
        Lunit-DINO & 95.4$\pm$0.5 & 95.0$\pm$0.6 & 93.3$\pm$3.1 & 86.6$\pm$8.3 & 61.6$\pm$8.8 & 88.1$\pm$1.4 & 85.5$\pm$0.3 & 80.1$\pm$2.6\\
        Lunit-BT\footnote[1] & 94.3$\pm$0.3 & 94.7$\pm$0.3 & 94.4$\pm$0.3 & 94.1$\pm$0.4 & 92.3$\pm$0.8 & \textbf{91.0$\pm$0.2} & \textbf{90.7$\pm$0.2} & \textbf{89.2$\pm$0.1}\\
        CTransPath\footnote[1] & \textbf{97.4$\pm$0.2} & \textbf{97.4$\pm$0.2} & \textbf{97.4$\pm$0.2} & \textbf{96.9$\pm$0.1} & \textbf{95.1$\pm$0.3} & 89.0$\pm$0.5 & 87.0$\pm$0.8 & 85.2$\pm$1.5\\
        PathoDuet\footnote[1] & 96.1$\pm$0.4 & 96.3$\pm$0.2 & 95.9$\pm$0.3 & 95.5$\pm$0.2 & 93.3$\pm$0.8 & 87.1$\pm$0.5 & 87.0$\pm$0.2 & 85.8$\pm$0.4\\
        iBOT & 93.3$\pm$0.5& 90.1$\pm$1.2& 84.6$\pm$4.4& 78.3$\pm$6.2& 45.1$\pm$10.2& 83.2$\pm$0.5& 77.8$\pm$1.9& 68.9$\pm$1.9 \\
        MoCo\footnote[1] & 93.2$\pm$0.5& 93.0$\pm$1.1& 90.8$\pm$1.4& 88.8$\pm$0.4& 86.0$\pm$0.4& 81.2$\pm$0.3& 80.1$\pm$0.4& 78.6$\pm$0.5 \\
        \hline
    \end{tabular}
    
    \vspace{2em}
    
    \begin{tabular}{||C{1.7cm}||C{1.32cm}C{1.32cm}C{1.32cm}C{1.32cm}|C{1.15cm}C{1.1cm}C{1.3cm}||}   
        \hline
        \multirow{2}{*}{Method}  &  \multicolumn{4}{c|}{BACH} & 
        \multicolumn{3}{c||}{MHIST}\\
        & $\eta=0$ & $\eta=0.2$ & $\eta=0.4$ & $\eta=0.6$ &
        $\eta=0$ & $\eta=0.2$ & $\eta=0.4$ \\
        
        \hline
        \hline
        Baseline & 78.2$\pm$2.1 & 65.1$\pm$3.4 & 41.3$\pm$6.9 & 40.5$\pm$6.3 & 82.9$\pm$0.9 & 72.1$\pm$3.1 & 60.2$\pm$0.7\\
        GCE & 76.9$\pm$0.2 & 68.5$\pm$1.8 & 61.3$\pm$1.5 & 52.3$\pm$2.7& 84.3$\pm$0.6 & 76.9$\pm$0.6 &  59.2$\pm$6.7\\
        APL & 77.5$\pm$0.6 & 73.1$\pm$0.6 & 64.6$\pm$2.3 & 51.7$\pm$4.9 & 83.8$\pm$0.9 & 72.1$\pm$8.9 & 68.2$\pm$0.6 \\

        DivideMix & 79.6$\pm$0.1 & 73.8$\pm$2.0 & 64.9$\pm$3.1 & 50.9$\pm$2.1 & 84.2$\pm$0.2 & \textbf{80.9$\pm$0.4} & 72.2$\pm$1.4\\
        \hline
        Phikon & 77.4$\pm$1.3 & 69.3$\pm$2.2 & 55.1$\pm$9.7 & 40.0$\pm$4.8 & \textbf{84.5$\pm$0.5} & 75.0$\pm$0.9 & 57.9$\pm$4.8\\
        RetCCL\footnote[1] &  73.5$\pm$0.5& 71.7$\pm$0.3 & 68.7$\pm$1.7 & 61.0$\pm$1.3 & 81.2$\pm$0.7 & 78.1$\pm$1.2 &73.0$\pm$3.5 \\
        Lunit-Dino & 80.7$\pm$0.1 &  72.8$\pm$2.1&58.7$\pm$3.6 & 41.0$\pm$3.5 & 79.1$\pm$2.1 & 75.1$\pm$1.8 & 64.3$\pm$4.1\\
        Lunit-BT\footnote[1] & 81.2$\pm$1.3 & 78.7$\pm$0.9 & 74.3$\pm$1.5 & 65.7$\pm$1.9 & 79.6$\pm$0.4 & 78.1$\pm$1.5 & \textbf{76.0$\pm$0.8}\\
        CTransPath\footnote[1] & \textbf{83.7$\pm$0.6} & \textbf{80.4$\pm$1.6} & \textbf{77.2$\pm$1.3} & \textbf{72.3$\pm$1.0} & 80.4$\pm$0.6 & 79.6$\pm$0.7 & 75.1$\pm$1.2\\
        PathoDuet\footnote[1] & 79.0$\pm$2.5 & 75.4$\pm$2.1 & 72.7$\pm$2.2 & 68.0$\pm$1.8 & 78.3$\pm$1.0 & 76.4$\pm$0.9 & 71.1$\pm$3.5\\
        iBOT & 79.2$\pm$2.0& 62.7$\pm$5.2& 51.2$\pm$4.7& 34.0$\pm$7.7& 82.1$\pm$0.7 &72.3$\pm$1.9& 56.8$\pm$1.2\\
        MoCo\footnote[1] & 71.2$\pm$0.5& 68.8$\pm$1.2& 65.6$\pm$1.4& 58.2$\pm$1.5& 78.2$\pm$0.5& 76.1$\pm$1.2& 72.6$\pm$1.4\\
        \hline
    \end{tabular}
    
    \vspace{2em}
    
    \begin{tabular}{||C{1.7cm}||C{1.32cm}C{1.32cm}C{1.32cm}C{1.32cm}|C{1.15cm}C{1.1cm}C{1.3cm}||} 
        \hline
        \multirow{2}{*}{Method} & \multicolumn{4}{c|}{LC25000}& 
        \multicolumn{3}{c||}{GasHisSDB} \\
        & $\eta=0$ & $\eta=0.2$ & $\eta=0.4$ & $\eta=0.6$ & $\eta=0$ & $\eta=0.2$ & $\eta=0.4$\\
        \hline
        \hline
        Baseline & \textbf{100$\pm$0.0} & 95.0$\pm$0.4 & 77.5$\pm$3.7 & 75.8$\pm$4.2  & 98.2$\pm$0.4& 87.7$\pm$0.2 & 65.5$\pm$2.4\\
        GCE & 99.9$\pm$0.1  & \textbf{99.8$\pm$0.1} & 99.4$\pm$0.2 & 96.9$\pm$0.2 & 98.3$\pm$0.1& 96.0$\pm$0.2 & 74.9$\pm$5.0\\
        APL & 99.9$\pm$0.1 & 99.7$\pm$0.1 & 99.4$\pm$0.1 & 98.1$\pm$0.2 & 98.1$\pm$0.1& 96.5$\pm$0.1 & 75.4$\pm$0.5\\
        DivideMix & 99.7$\pm$0.2 & 99.6$\pm$0.1 & 99.4$\pm$0.2 & 98.4$\pm$0.1 &97.9$\pm$0.1 & \textbf{97.3$\pm$0.1}& 93.5$\pm$0.4\\
        \hline
        Phikon & 99.9$\pm$0.1 & 99.6$\pm$0.1 & 99.1$\pm$0.1 & 97.7$\pm$0.5  & \textbf{99.6$\pm$0.1}& 86.7$\pm$1.5 & 64.9$\pm$1.6 \\
        RetCCL\footnote[1] & 98.7$\pm$0.2 & 97.8$\pm$0.2 & 97.6$\pm$0.2 & 97.0$\pm$0.2  & 97.2$\pm$0.1 & 95.3$\pm$0.5 & 87.3$\pm$2.5\\
        Lunit-DINO & 99.6$\pm$0.2 & 99.1$\pm$0.1 & 99.0$\pm$0.1 & 97.2$\pm$0.6  & 99.2$\pm$0.1 & 84.9$\pm$2.1 & 65.7$\pm$2.2\\
        Lunit-BT\footnote[1] & 99.9$\pm$0.1 & 99.6$\pm$0.3 & 99.1$\pm$0.1 & 98.6$\pm$0.1  & 98.5$\pm$0.1 & 97.2$\pm$0.3 & 91.6$\pm$3.2\\
        CTransPath\footnote[1] & 99.9$\pm$0.1 & 99.3$\pm$0.2 & 99.1$\pm$0.1 & \textbf{98.9$\pm$0.1} & 98.7$\pm$0.1 & \textbf{97.3$\pm$0.2} & \textbf{94.6$\pm$0.3}\\
        PathoDuet\footnote[1] & 99.9$\pm$0.1 & \textbf{99.8$\pm$0.1} & \textbf{99.6$\pm$0.1} & 98.5$\pm$0.1  & 98.3$\pm$0.1 & 95.9$\pm$0.9 & 92.4$\pm$1.1\\
        iBOT & 99.9$\pm$0.1& 95.7$\pm$1.3& 89.2$\pm$0.7& 67.8$\pm$8.2& 97.3$\pm$0.4& 80.1$\pm$1.2& 60.6$\pm$3.8\\
        MoCo\footnote[1] & 98.5$\pm$0.1& 98.2$\pm$0.1& 97.9$\pm$0.1& 97.4$\pm$0.2& 96.2$\pm$0.2& 91.0$\pm$0.4&85.9$\pm$0.6 \\
        \hline
    \end{tabular}
    
    \label{tab:sym}
\end{table}

\begin{figure}[h]
\centering
\includegraphics[width=12.5cm]{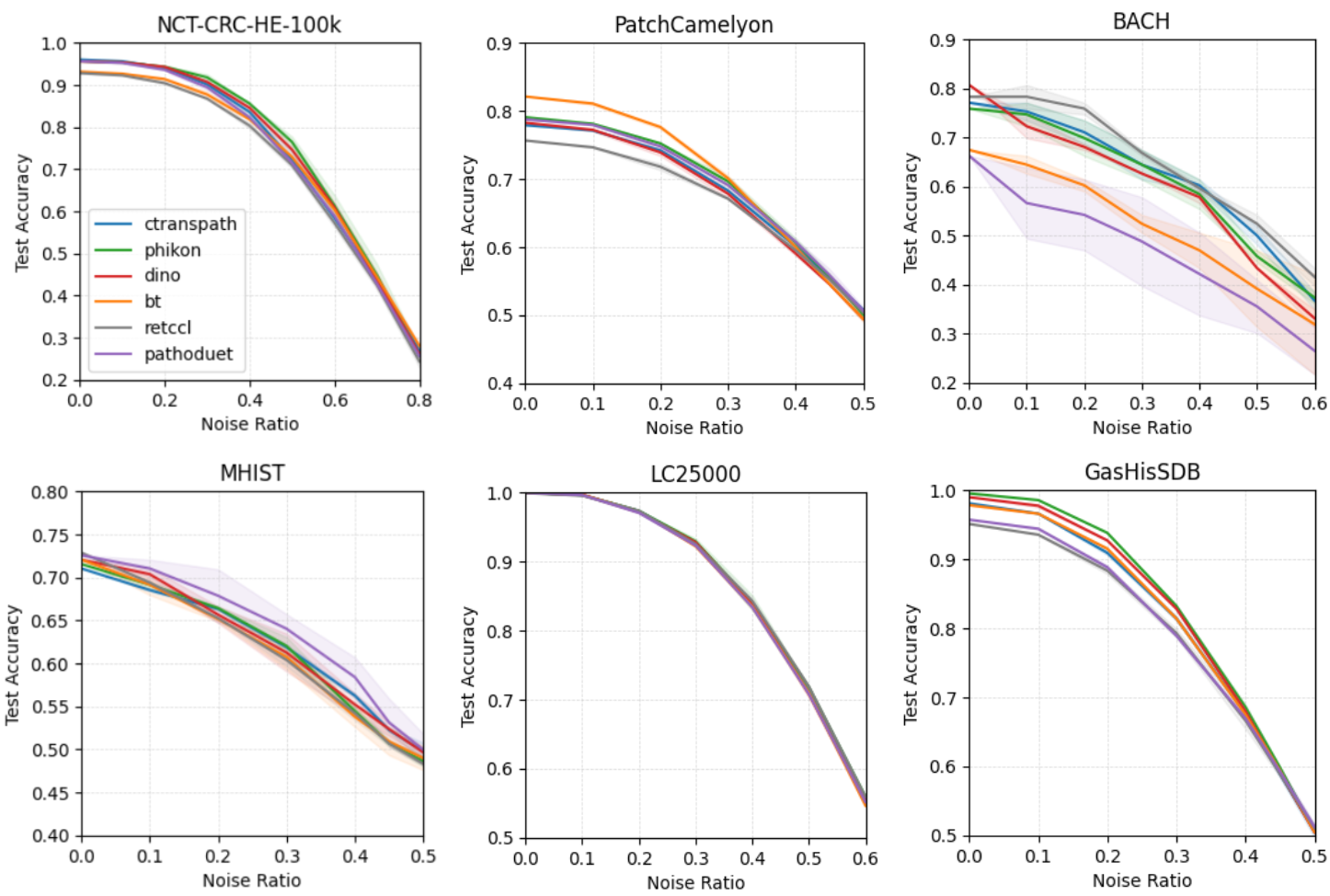}
\caption{Average test accuracies (4 runs) of k-NN classifiers (k=5) trained on 10\% of train datasets over different label noise ratios. Shaded areas represent standard deviation.}
\label{fig:knn}
\end{figure}

\begin{figure}[h]
\centering
\includegraphics[width=12.cm]{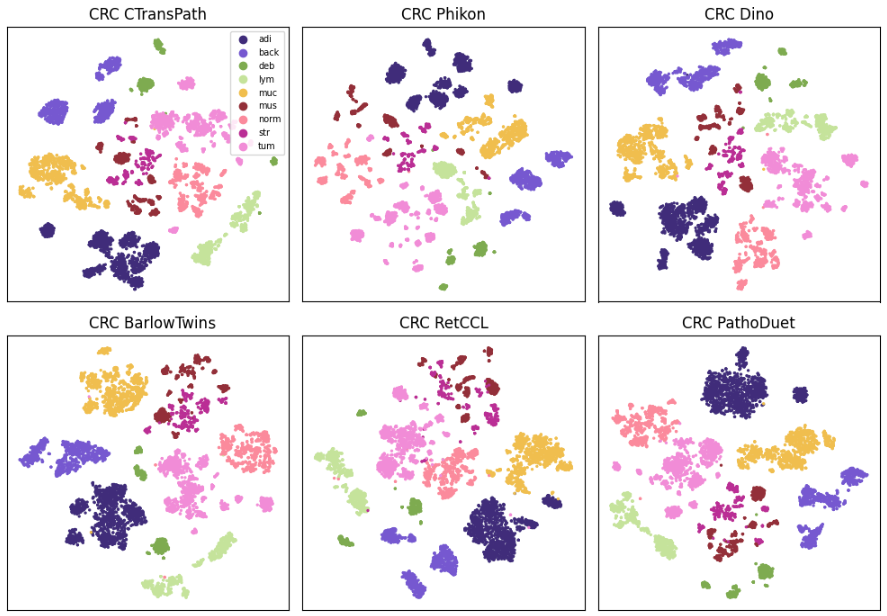}
\caption{t-SNE representations of NCT-CRC-HE-100k deep embeddings extracted from various histopathology foundation 
            models.}
\label{fig:tsne}
\end{figure}

One might think that these two models extract inferior representations compared to others. However, we demonstrate that this is not true through a few-shot learning experiment using k-nearest neighbors (k-NN) classifiers trained on 10\% of each dataset. Idea is to test the inherent resistance of embeddings and demonstrate that, without proper training of a linear head, they exhibit uniform reactions to label noise. Results, illustrated in \figureref{fig:knn}, confirm that hypothesis. Furthermore, upon closer examination through t-SNE visualization of NCT-CRC-HE-100k in \figureref{fig:tsne} it becomes evident that no model extracts a worse representation. Thus the observed performance differences are not due to the quality of the learned representations, but rather to the noise-resilient property leveraged by the linear classifier when trained with contrastive embeddings.

That explains why in \figureref{fig:curves}, at $\eta = 0$, Phikon and Lunit-DINO perform really well but, at higher noise rates, some contrastive methods start with lower accuracies than them at $\eta = 0$, yet subsequently outpace them, thereby showcasing their enhanced resilience to label noise. A particularly prominent example is PatchCamelyon with RetCCL, where the test accuracy starts at 85.8\%, while Phikon and Lunit-DINO start at 86.4\% and 88.1\%. However, RetCCL maintains stability as $\eta$ increases, while the other two decline. 
Under asymmetric noise, the conclusion is consistent with the previous observations as shown in \tableref{tab:asym}. Specifically, on the three multiclass datasets, training with contrastive deep embeddings yields the best results. Moreover, the results on the two ImageNet backbones under both noise types further support our findings, with iBOT exhibiting a consistently steeper decline in performance than MoCo as the noise rate increases.

\begin{table}[h]
    \tiny
    \centering
    \caption[]{Average test accuracies with standard deviation on NCT-CRC-HE-100k, BACH and LC25000 under different asymetric label noise ratios (\%, 2 runs for image methods, 4 runs for deep embedding methods). Class transition matrices are given in appendix \ref{asym}. \footnote[1] :Foundation model trained with contrastive learning.}
    {\setlength{\tabcolsep}{0.95em}
    \begin{tabular}{||c||cc|cc|cc||}
        \hline
        \multirow{2}{*}{Method} & \multicolumn{2}{c|}{NCT-CRC-HE-100k}& 
        \multicolumn{2}{c|}{BACH}&
        \multicolumn{2}{c||}{LC25000}\\
        & $\eta=0.2$ & $\eta=0.4$ & $\eta=0.2$ & $\eta=0.4$ & $\eta=0.2$ & $\eta=0.4$\\
        \hline
        \hline
        Baseline & 88.6$\pm$0.9& 78.4$\pm$1.3& 60.4$\pm$4.1& 40.2$\pm$8.4& 95.1$\pm$1.1& 69.3$\pm$1.2\\
        GCE& 94.5$\pm$0.3& 91.3$\pm$1.7& 65.4$\pm$1.3& 56.8$\pm$2.7& 97.2$\pm$0.1& 92.7$\pm$0.4\\
        APL & 94.5$\pm$0.4& 92.0$\pm$0.3& 70.5$\pm$0.9& 60.7$\pm$2.2& 97.2$\pm$0.3& 93.4$\pm$0.5\\
        DivideMix & 96.0$\pm$0.1& 95.4$\pm$0.2& 72.4$\pm$1.5& 61.3$\pm$2.0& 98.9$\pm$0.1& 96.5$\pm$0.2\\
        \hline
        Phikon & 88.0$\pm$4.1& 74.9$\pm$7.8& 66.9$\pm$3.5& 46.7$\pm$3.6& 98.9$\pm$0.1& 96.3$\pm$0.7\\
        RetCCL\footnote[1]& 93.9$\pm0.4$& 93.4$\pm$0.3& 77.4$\pm$1.6& 61.4$\pm$5.2& 97.6$\pm$0.1& 96.8$\pm$0.2\\
        Lunit-DINO & 93.1$\pm$2.6& 85.7$\pm$6.4& 65.1$\pm$4.1& 51.2$\pm$3.9& 99.1$\pm$0.1& 97.3$\pm$0.2\\
        Lunit-BT\footnote[1] & 94.5$\pm$0.2& 94.2$\pm$0.7& 69.3$\pm$4.0& 57.5$\pm$8.4& 99.5$\pm$0.1& 97.9$\pm$0.3\\
        CTransPath\footnote[1] & \textbf{97.4$\pm$0.2}& \textbf{97.1$\pm$0.3}& \textbf{79.0$\pm$1.3}& \textbf{66.9$\pm$14.7}& 99.2$\pm$0.3& \textbf{98.5$\pm$0.1}\\
        PathoDuet\footnote[1]& 95.8$\pm$0.3& 95.6$\pm$0.4& 73.1$\pm$0.8& 62.2$\pm$3.5& \textbf{99.6$\pm$0.1}& 98.1$\pm$0.7\\
        iBOT & 82.4$\pm$3.4& 66.9$\pm$8.7& 58.6$\pm$4.6& 45.8$\pm$3.5& 93.1$\pm$3.3& 77.5$\pm$3.1\\
        MoCo\footnote[1] & 92.9$\pm$0.3& 91.2$\pm$0.4& 66.8$\pm$0.6& 62.3$\pm$1.4& 92.5$\pm$4.2& 88.3$\pm$2.4 \\
        \hline
    \end{tabular}
    }
    \label{tab:asym}
\end{table}
\section{Limitations}
Although the study shows that contrastive learning effectively mitigates the label noise challenge, it does not completely eliminate it. Specifically, for relatively small datasets such as BACH and MHIST, the stability over varying noise rates is lower than the one observed with larger datasets. Similarly, asymmetric noise appears to be more challenging to resist. Thus, there is still a need for further research to develop methods that can completely overcome the label noise issue. Potential avenues for future research include exploring the combination of contrastive embeddings with other noise resilience methods such as label cleaning or robust losses or investigating the impact of fine-tuning the non-contrastive backbones with contrastive learning.
For control over noise rates we used synthetic label noise over public datasets. A validation study using real-world noisy datasets could conclude into the effectiveness of contrastive embeddings in handling label noise.

\section{Conclusion}
Our study highlights the effectiveness of contrastive-based deep embedding training in bolstering the robustness of histopathology classifiers under noisy labels. Through extensive experiments across multiple benchmark datasets, we showcase the superior noise resilience of linear classifiers trained with contrastive-based deep embeddings when compared to image-based methods and non-contrastive embeddings. By elucidating these findings, our aim is to offer valuable insights and guidelines for the future development of histopathological foundation models. Notably, the majority of current histology foundation models overlook the assessment of label noise resilience in their performance evaluations, despite it being one of the most significant challenges in digital histopathology.

\midlacknowledgments{This work was supported by Primaa.}

\bibliography{biblio}

\appendix

\vspace{10em}

\section{Experimental Settings}\label{settings}
 As a baseline, we use a ResNet-50 model trained with the CCE loss starting from ImageNet \cite{imagenet} weights.  For baseline runs we used a stochastic gradient descent (SGD) optimizer \cite{sgd} with learning rate of 0.0005 coupled to cosine annealing scheduler \cite{cosinescheduler} and early stopping with 20 patience epochs. For GCE, we set $q$ to $0.7$ and used the same settings as baseline. For APL, we used the normalized cross entropy - reverse cross entropy (NCE-RCE) loss with $\alpha = 0.6$ and $\beta=0.4$. For both GCE and APL we did a five epochs warmup with CCE loss. For DivideMix, we used default settings and a learning rate of 0.001 for 50 epochs. For embedding-based trainings, a four-layer linear neural network serves as the classifier model. We used grid search to find out optimal learning rate, batch size, patience and Gaussian noise $\sigma$ for each backbone-dataset combinations. All runs are conducted on Nvidia GeForce RTX 2080 Ti GPUs.

\section{Datasets Details}\label{dataset}

 NCT-CRC-HE-100K consists of 100,000 hematoxylin and eosin (H\&E) stained 224x224 histological images of human colorectal cancer (CRC) divided in 9 tissue classes (adipose, background, debris, lymphocytes, mucus, smooth muscle, normal colon mucosa, cancer-associated stroma, colorectal adenocarcinoma epithelium). The test set uses a different CRC-VAL-HE-7K dataset consisting of 7,180 patches featuring all nine classes. PatchCamelyon binary dataset contains 327,680 stained 96x96 patches derived from histopathological scans of lymph node sections with positive label indicating presence of metastatic tissue. It provides train, validation and test sets. PatchCamelyon is derived from the Camelyon16 challenge \cite{camelyon16}. BACH dataset is composed of H\&E stained breast histology microscopy 2048x1536 patches. Train and test sets respectively possess 400 and 83 patches distributed in 4 classes according to the predominant cancer type in each image (normal, benign, \textit{in situ} carcinoma, invasive carcinoma). As the ground truth was not available for the test set, we asked a breast pathologist expert to label the 100 patches. Among the 100, 17 were considered uncertain due to a lack of contextual information on the patch so we chose to not use these 17 patches in order to have a clean ground truth. MHIST is made up of 3,152 H\&E stained 224x224 images of colorectal polyps split into 2175 and 977 for train/test sets. Binary classes, hyperplastic polyp (HP) or sessile serrated adenoma (SSA), indicate the predominant histological pattern. LC25000 dataset contains five classes of 5,000 768x768 images of lung and colon tissues. The classes are: colon adenocarcinomas, benign colonic tissues, lung adenocarcinomas, lung squamous cell carcinomas and benign lung tissues. GasHisSDB consists of 32,284 160x160 H\&E stained gastric cancer pathology patches. Classes are abnormal and normal. In these two last datasets there is no test set provided so we split ourselves train set into 80:20 for train/test sets then 80:20 for train/validation sets.
 
For image training methods, we employ a traditional preprocessing and augmentation pipeline. For the training set, we incorporate resizing, horizontal and vertical flipping, color jittering, and normalization. For the validation and test sets, we simplify the process by only resizing and normalizing the images. The resizing dimensions are set to 512x512 for BACH and LC25000 datasets, while for the remaining datasets, we maintain the original sizes.

For embedding extractions, we consider the fact that backbones require an input size of 224x224. Therefore, for oversized images, such as those in the BACH and LC25000 datasets, we resize them to this resolution. For undersized images, like those in the PatchCamelyon and GasHisSDB datasets, we experiment with both resizing and padding. Our results indicate that resizing yields better outcomes. The size of the embeddings varies depending on the backbone used. For Phikon, CTransPath, and PathoDuet, the embedding size is 768, while for RetCCL and Lunit-BT, it is 2048. Lunit-DINO, on the other hand, uses an embedding size of 384.

For training methods based on embeddings, we introduce an additional augmentation to the training set by adding random Gaussian noise, scaled with $\sigma$.

\vspace{8em}

\section{Asymetric Noise Transition Matrices}\label{asym}
\begin{table}[h]
    \centering
    \caption{NCT-CRC-HE-100k class transition matrix.}
    \begin{tabular}{|c|c|c|c|c|c|c|c|c|c|}
        \hline
         & ADI & BACK & DEB & LYM & MUC & MUS & NORM & STR & TUM \\
        \hline
        ADI & $1-\eta$ & $0$ & $0$ & $0$ & $0$ & $\frac{\eta}{2}$ & $\frac{\eta}{2}$ & $0$ &0\\
        \hline
        BACK & $0$ & $1-\eta$ & $\frac{\eta}{3}$ & $\frac{\eta}{3}$ & $\frac{\eta}{3}$ & $0$ & $0$ & $0$&0 \\
        \hline
        DEB & $0$ & $\frac{\eta}{3}$ & $1-\eta$ & $\frac{\eta}{3}$ & $\frac{\eta}{3}$ & $0$ & $0$ & $0$ &0\\
        \hline
        LYM & $0$ & $\frac{\eta}{3}$ & $\frac{\eta}{3}$ & $1-\eta$ & $\frac{\eta}{3}$ & $0$ & $0$ & $0$ &0\\
        \hline
        MUC & $0$ & $\frac{\eta}{3}$ & $\frac{\eta}{3}$ & $\frac{\eta}{3}$ & $1-\eta$ & $0$ & $0$ & $0$ &0\\
        \hline
        MUS & $\frac{\eta}{2}$ & $0$ & $0$ & $0$ & $0$ & $1-\eta$ & $\frac{\eta}{2}$ & $0$ &0\\
        \hline
        NORM & $\frac{\eta}{2}$ & $0$ & $0$ & $0$ & $0$ & $\frac{\eta}{2}$ & $1-\eta$ & $0$&0 \\
        \hline
        STR & $0$ & $0$ & $0$ & $0$ & $0$ & $0$ &0 & $1-\eta$ & $\eta$\\
        \hline
        TUM & 0& $0$ & $0$ & $0$ & $0$ & $0$ & $0$ & $\eta$ & $1-\eta$ \\
        \hline
    \end{tabular}
\end{table}
\begin{table}[h]
    \centering
    \caption{BACH class transition matrix.}
    \begin{tabular}{|c|c|c|c|c|}
        \hline
         & Benign & CIS & CI & Normal \\
        \hline
        Benign & $1-\eta$ & $\eta$ & $0$ & $0$ \\
        \hline
        CIS & $\frac{\eta}{2}$ & $1-\eta$ & $\frac{\eta}{2}$ & $0$ \\
        \hline
        CI & $0$ & $\frac{\eta}{2}$ & $1-\eta$ & $\frac{\eta}{2}$ \\
        \hline
        Normal & $0$ & $0$ & $\eta$ & $1-\eta$ \\
        \hline
    \end{tabular}
\end{table}
\begin{table}[h]
    \centering
    \caption{LC25000 class transition matrix.}
    \begin{tabular}{|c|c|c|c|c|c|}
        \hline
        & Colon ACA & Benign Colon & Lung ACA & Benign Lung& Lung SCC\\
        \hline
        Colon ACA & $1-\eta$ & $\frac{\eta}{2}$ & $\frac{\eta}{2}$ & $0$ & $0$ \\
        \hline
        Benign Colon & $\frac{\eta}{2}$ & $1-\eta$ & $0$ & $\frac{\eta}{2}$ & $0$ \\
        \hline
        Lung ACA & $\frac{\eta}{2}$ & $0$ & $1-\eta$ & $0$ & $\frac{\eta}{2}$ \\
        \hline
        Benign Lung & 0 & $\eta$ & $0$ & $1-\eta$ & $0$ \\
        \hline
        Lung SCC & $0$ & $0$ & $\eta$ & $0$ & $1-\eta$ \\
        \hline
    \end{tabular}
\end{table}

\end{document}